\documentclass[runningheads]{llncs}
\usepackage{graphicx}

\usepackage{tikz}
\usepackage{comment}
\usepackage{amsmath,amssymb} 
\usepackage{color}

\usepackage[accsupp]{axessibility}  

\usepackage{multirow}
\usepackage{csquotes}

\usepackage{amsmath,amsfonts,bm}

\def\eqref#1{equation~\ref{#1}}

\def\1{\bm{1}}

\newcommand{\test}{\mathcal{D_{\mathrm{test}}}}

\def\vd{{\bm{d}}}

\def\vp{{\bm{p}}}

\DeclareMathAlphabet{\mathsfit}{\encodingdefault}{\sfdefault}{m}{sl}
\SetMathAlphabet{\mathsfit}{bold}{\encodingdefault}{\sfdefault}{bx}{n}
\newcommand{\tens}[1]{\bm{\mathsfit{#1}}}

\def\tD{{\tens{D}}}

\def\tI{{\tens{I}}}

\def\tK{{\tens{K}}}

\def\tM{{\tens{M}}}

\def\tO{{\tens{O}}}

\def\tU{{\tens{U}}}
\def\tV{{\tens{V}}}
\def\tW{{\tens{W}}}

\usepackage{enumitem}
\usepackage[super]{nth}

\newcommand\inlineeqno{\stepcounter{equation}\ (\theequation)}

\usepackage{xcolor,pifont}
\newcommand*\colourcheck[1]{%
  \expandafter\newcommand\csname #1check\endcsname{\textcolor{#1}{\ding{52}}}%
}
\colourcheck{blue}
\colourcheck{green}
\colourcheck{red}
\colourcheck{black}

\usepackage{colortbl}

\definecolor{c_muns}{rgb}{0.88,1,1}
\definecolor{c_msup}{rgb}{1.0,0.88,1}
\definecolor{c_data}{rgb}{0.9,0.9,0.9}
\definecolor{c_lowbest}{rgb}{1.0,0.7,0.7}
\definecolor{c_highbest}{rgb}{0.7,0.7,1.0}

\newcommand{\etal}{\textit{et al. }}

\begin{document}
\pagestyle{headings}
\mainmatter
\def\ECCVSubNumber{7689}  


\title{Positional Information is All You Need: A Novel Pipeline for Self-Supervised SVDE from Videos} 

\titlerunning{Positional Information Is All You Need}
%
\author{Juan Luis Gonzalez Bello \and
Jaeho Moon \and
Munchurl Kim}
\authorrunning{J.L. Gonzalez et al.}
%

\institute{Korea Advanced Institute of Science and Technology\\
\email{juanluisgb@kaist.ac.kr} \email{james16@kaist.ac.kr} \email{mkimee@kaist.ac.kr}
}
\maketitle

\begin{figure}
  \centering 
  \vspace*{-5mm}
  \includegraphics[width=1.0\textwidth]{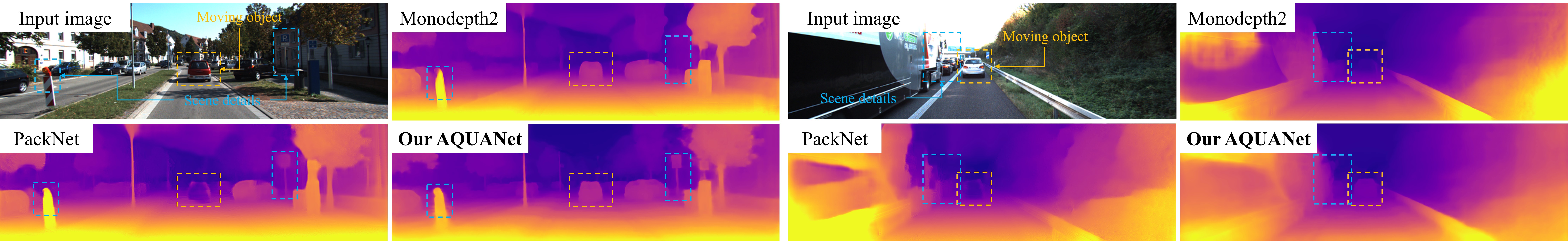}
  \vspace*{-6mm}
  \caption{Our proposed method consistently learns more details and holes-free dense depths from monocular videos.}
  \label{fig:opening_img}
  \vspace*{-10mm}
\end{figure}

\begin{abstract}
    Recently, much attention has been drawn to learning the underlying 3D structures of a scene from monocular videos in a fully self-supervised fashion. One of the most challenging aspects of this task is handling the independently moving objects as they break the rigid-scene assumption. For the first time, we show that pixel positional information can be exploited to learn SVDE (Single View Depth Estimation) from videos. Our proposed moving object (MO) masks, which are induced by shifted \textit{positional information} (SPI) and referred to as `SPIMO' masks, are very robust and consistently remove the independently moving objects in the scenes, allowing for better learning of SVDE from videos. 
    Additionally, we introduce a new adaptive quantization scheme that assigns the best per-pixel quantization curve for our depth discretization. Finally, we employ existing boosting techniques in a new way to further self-supervise the depth of the moving objects. With these features, our pipeline is robust against moving objects and generalizes well to high-resolution images, even when trained with small patches, yielding state-of-the-art (SOTA) results with almost 8.5$\times$ fewer parameters than the previous works that learn from videos. We present extensive experiments on KITTI and CityScapes that show the effectiveness of our method. 
\end{abstract}

\section{Introduction}

\begin{figure*}[t]
  \centering 
  \includegraphics[width=0.99\textwidth]{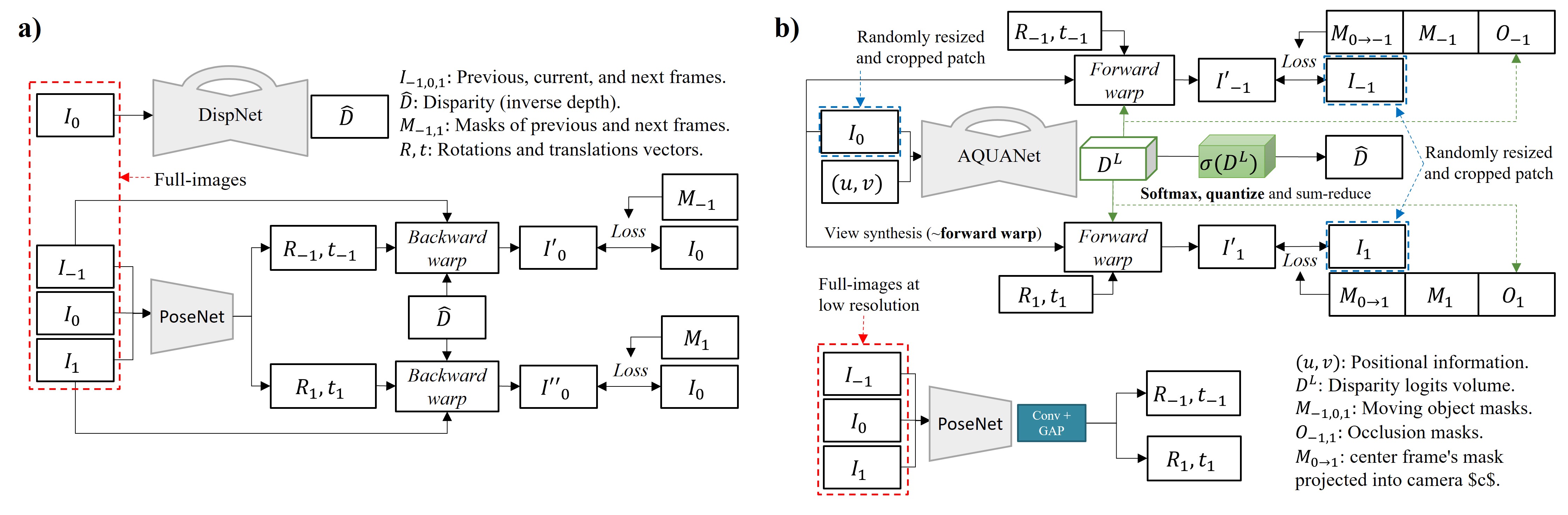}
  \vspace*{-4mm}
  \caption{Traditional (\textbf{a}) and proposed (\textbf{b}) pipelines for learning SVDE from videos.}
  \label{fig:pipelines}
  \vspace*{-4mm}
\end{figure*}

Estimating depth from a single image has multiple applications ranging from robotics, navigation, and computational imaging. However, learning single view depth estimation (SVDE) from monocular videos is challenging as camera poses and depths are estimated up to an unknown and inconsistent scale, making the training of deep neural networks unstable at the early epochs. In addition, objects moving at the same speed as the camera will have no disparities thus will be assigned infinite depths. On the other hand, objects moving faster than the camera will have larger disparities, resulting in implausibly closer depths. In general, dynamically moving objects are likely to cause incorrect depths along their boundaries due to their weak or in-existent 3D self-supervision.

Traditionally, deep-learning-based pipelines \cite{sfmlearner,monodepth2,packing3d} for learning SVDE from videos rely on: (i) a CNN that directly estimates inverse depth; (ii) a CNN that directly regresses relative camera poses; and (iii) a backward warping-based loss function that measures the similarity between the target center frame and the warped reference frames. Independently moving objects in such loss functions are commonly removed by moving object masks that are computed by analyzing the reconstruction errors between the target frame and the multiple warped reference frames \cite{monodepth2}. 
In contrast, we present a novel learning pipeline that relies on pixel positional information to compute the moving object masks and forward warping-based loss functions (image synthesis) for learning accurate SVDE from high-resolution monocular videos. We summarize our contributions as:
\begin{itemize}[leftmargin=*,noitemsep,topsep=0.5pt]
\vspace*{0.5mm}
\item \textbf{Robust estimation of moving object masks by shifted positional information.} To ignore independently moving objects (that can break the rigid-scene assumption) we propose novel moving object masks induced by shifted \textit{positional information}, referred to as SPIMO masks. These masks are more robust and less sparse than the traditional auto-masks \cite{monodepth2}, allowing for better learning of SVDE from monocular videos. In addition, SPIMO masks do not require any additional data such as segmentation masks or optical flow nor domain specific assumptions. Hence \textit{positional information is all you need}.

\vspace*{0.5mm}
\item \textbf{Depth cue-preserving data augmentation.} For the first time, we show that encouraging a network to generate disparity estimates with values that scale in proportion to the image resolutions, when training with random resize and crop, benefits the learning process as it exploits one of the best known depths cues, relative object size - \textit{the larger the objects, the closer they are.} 


%

\vspace*{0.5mm}
\item \textbf{Adaptive disparity quantization.} Disparity / depth discretization converts the harder regression task into an easier classification task. However, in contrast with previous works that use linear or exponential quantization \cite{deep3d,falnet,pladenet} for stereoscopic view synthesis, we propose a novel adaptive quantization scheme for free view synthesis. In our adaptive quantization scheme, the proposed adaptive quantization network (AQUANet) automatically learns to assign a quantization curve that dedicates more distant sampling levels if the target pixel is likely to be far, and a quantization curve with more close-by sampling levels if it is closer to the camera, improving the accuracy of the final aggregated depth map.




\end{itemize}




\section{Related Work}

\subsubsection{Fully Supervised Learning of SVDE.}
The early fully-supervised methods for SVDE utilized deep neural networks to perform depth map regression. For example, in \cite{eigen}, one network estimated global depth predictions, and the other network refined the predictions with local context. 
To address the slow convergence of regression models, discretizing depth values allowed the SVDE task to be an ordinal regression problem \cite{dorn}. To further improve the quality of high-frequency details in SVDE, a double-estimation method for merging depth estimations at different resolutions was proposed in \cite{boosting_midas}. The resulting depth maps contained high-frequency details of the image while preserving the scene's structural consistency.
While the fully-supervised learning of the SVDE task has been gaining lots of attention, both the tasks of depth estimation and completion are explored with \cite{sparse_auxiliary}.
Although many methods are proposed for fully supervised learning of SVDE, the burden of acquiring the depth ground truths limits the scalability of the training dataset and the generalization of models.

\vspace*{-2mm}
\subsubsection{Self-Supervised Learning from Stereo.}
Without the depth ground truths, a neural network can learn to estimate depth by minimizing a photometric loss between the right (or left) view and a synthetic view when the training data consists of synchronized stereo images. 
To further supervise a CNN for SVDE from stereo, semi-global matching (SGM\cite{sgm0}) pseudo-ground have also been explored  \cite{depth_hints,infuse_classic}. 
In recent works  \cite{edge_of_depth}, segmentation masks have also been used in contour consistency losses to generate depth maps with more accurate borders.

The recent work \cite{falnet} proposed a multi-view occlusion module to account for the occluded regions on the reconstruction losses and an exponential disparity discretization which led to considerable performance improvements. In \cite{pladenet}, the authors introduced `neural positional encoding' (NPE) and a matting laplacian loss. NPE projects per-pixel positional information into a higher dimensional space via a fully-connected network to let the CNN reason about location-specific image characteristics (e.g., projection distortions, ground-vs-sky, etc.). Their matting laplacian loss guides the CNN to produce sharper depth boundaries by exploiting the sharpening properties of natural image matting \cite{matting}. 


\vspace*{-2mm}
\subsubsection{Self-Supervised Learning from Videos.}
Learning self-supervised SVDE from videos is a much more challenging task. While fixed stereo images are synchronized, video sequences have temporal distances and arbitrary camera poses between frames. This hinders the estimation of moving object depths and demands a secondary network for camera pose estimation as proposed by \cite{sfmlearner}. In \cite{sfmlearner}, `explainability'  masks were \textit{estimated} to handle such moving objects.

In Monodepth2 \cite{monodepth2}, the authors proposed auto-masking, which selects the pixels with the smallest projection errors to remove moving objects in the scene, improving SVDE performance. However, there is still a limitation on moving objects which move faster or slower than the camera. 
The recent work \cite{semguide} explored filtering out the sequences that have dynamic objects using a two-stage training strategy, semantics, and the PackNet \cite{packing3d} backbone. However, sub-sampling the training sequences can reduce the model's generalization capabilities.
Similarly, the work of \cite{sg_depth} utilized a semantic masking scheme by cross-domain training of semantic segmentation and SVDE.
Moreover, several methods \cite{scale_const_depth,semantic_aware_depth,nighttime,depth_all_day,feat_depth,hr_depth} have improved the depth estimation quality, but there is still a lot of room for improvement.
This paper suggests the SPIMO masks, which keep the model's scalability without requiring any additional data from different tasks. Using the SPIMO masks, we can successfully exclude moving objects for loss calculation and achieve outperforming results.

\section{Proposed Method}
We propose a novel pipeline for learning SVDE from monocular videos with SPIMO masks and adaptive disparity quantization, in which, \textit{positional information is all you need}. 

In the traditional pipeline \cite{monodepth2,packing3d,semguide} for learning SVDE from videos (depicted in Fig. \ref{fig:pipelines}-a) the reference previous and next frames ($\tI_{-1}$ and $\tI_1$) are projected onto a target frame ($I_0$) via backward warping. This projection is guided by the rigid flow that is produced by the target frame's inverse depth $\hat{\tD}$, the camera intrinsics, and the estimated camera extrinsics ($R_{-1,1},t_{-1,1}$). The projected images ($I'_0$ and $I''_0$) are then compared to the target frame with L1 and SSIM losses which simultaneously self-supervise the disparity and pose estimation networks. However, not all pixels in the projected images are useful for learning 3D geometries. Several factors such as illumination changes, homogeneous surfaces, reflections, occlusions, and moving objects can \enquote{distract} the networks from learning dense SVDE. To tackle this, previous works either learn to predict masks ($M_1$ and $M_{-1}$ in Fig. \ref{fig:pipelines}-a) that help the loss functions ignore such image regions \cite{sfmlearner}, or compute them by analyzing the photometric errors between the target frames and the reference frames \cite{monodepth2}. Note that the traditional pipeline requires full images to be fed into its depth and pose branches, thus quadratically increasing the training computational complexities with input resolutions.

Our new proposed pipeline is depicted in Fig. \ref{fig:pipelines}-b. The target frame ($I_0$) is projected via free-view synthesis (or forward warping) onto the reference frames ($I_{-1}$, $I_1$), generating new synthetic views seen from the reference camera positions. Forward warping is achieved by following the FAL-Net \cite{falnet}, which can be generalized to any camera position and augmented for \textit{adaptive} inverse depth discretization (see Section \ref{sec:formation_model} for more details). The synthetic views, denoted by $I'_{-1}$ and $I'_1$ in Fig. \ref{fig:pipelines}-b, are compared against the GT reference views with geometrically inspired reconstruction losses. Note that these losses not only consider masking the occluded contents in the target frame that are visible in the reference views but also the moving objects in the reference frames and the projected target frame, as denoted by the three masks in each reconstruction loss (e. g. $M_1$, $M_{0\rightarrow1}$, and $O_1$ for frame $I_{-1}$) in Fig. \ref{fig:pipelines}-b.

Similar to \cite{falnet}, our network indirectly learns SVDE by learning view synthesis. Remarkably, our pipeline admits randomly resized and cropped patches in its disparity/view synthesis branch while requiring low-res full images on its pose estimation branch (see Section \ref{sec:data_aug}), which allows it to learn depth from videos of arbitrary resolutions from small image patches.

Our training strategy is \textbf{two-staged}. Firstly, we let the SVDE network learn from all pixels (static and dynamic pixels) in the target and reference frames with an occlusion-aware synthesis loss following \cite{falnet}. Secondly, we devise a new method for detecting dynamically moving objects by measuring the dispersion of multiple depth estimates from this first stage network under several \textit{positional information perturbations} (see Section \ref{sec:moving_object}). In the second training stage, such robust moving object masks are used to effectively train an SVDE network from scratch, by ignoring the independently moving pixels in the scene in our new occlusion and moving-object-aware synthesis loss.
Additionally, we exploit recent advances in SVDE \cite{boosting_midas,dbooster} to self-supervise the moving object regions with \enquote{boosted} depth estimates, further improving our results (see Section \ref{sec:boosting}).

\begin{figure*}[t]
  \centering 
  \includegraphics[width=0.99\textwidth]{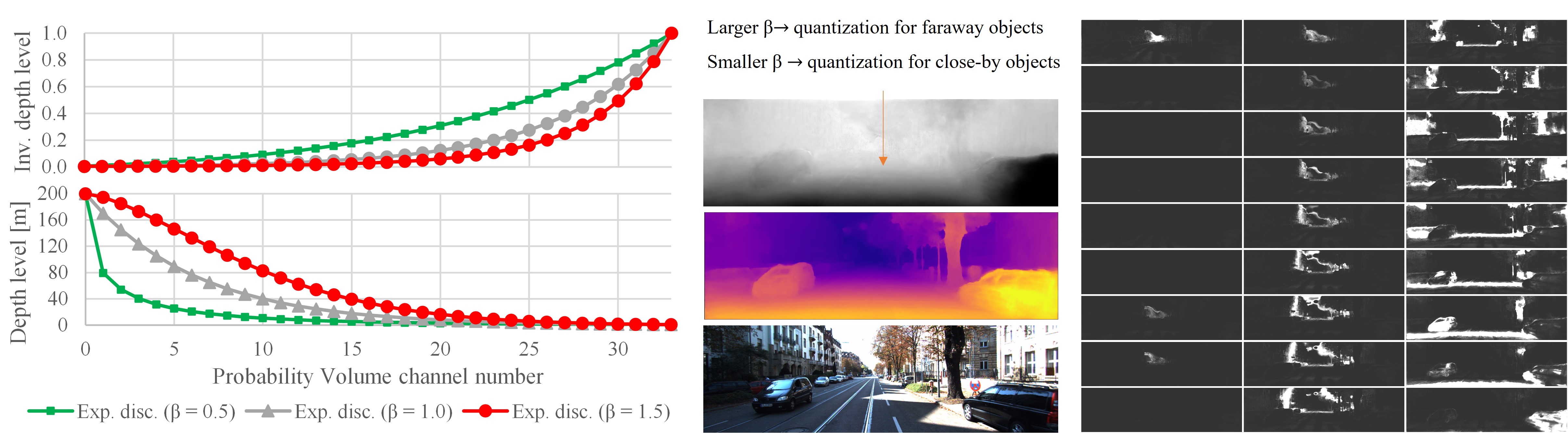}
  \vspace*{-3mm}
  \caption{Visualization of adaptive quantization. (Left) Inverse depth (disparity) quantization curves with adaptive parameter $\beta$. (Center) From bottom to top: Input image, disparity, and $\beta$ maps. (Right) Corresponding disparity probability volume.}
  \label{fig:adaptive_quant}
  \vspace*{-3mm}
\end{figure*}

\subsection{Adaptive Disparity Quantization for View Synthesis}
\label{sec:formation_model}
An approximation to forward warping an input image $\tI_0$ into a new camera position $c$ can be achieved by projecting $\tI_0$ into $c$ by $N$ depth planes and by summing all projected images weighted by their per-pixel depth probability distribution seen from $c$. We achieve this by generalizing and improving the synthesis method in \cite{falnet}, which is originally devised for fixed stereoscopic view synthesis, by incorporating camera intrinsics ($\tK_0$ and $\tK_c$) and extrinsics ($R_c$ and $t_c$) into the warping operation $g(\cdot)$. The synthetic view seen from camera position $c$ can then be expressed by
\begin{equation} \label{eq:synth_free}
\tI'_c = \textstyle \sum_{n=0}^{N-1} g\left(\tI_0, d_n,  R_c, t_c, \tK_0, \tK_c\right) \odot \tD^{P_{0 \rightarrow c}}_n,
\end{equation}
where $d_n$ is the inverse depth quantization level and $\odot$ denotes the Hadamard product. $\tD^{P_{0 \rightarrow c}}$ is the projected disparity probability volume seen from camera $c$, described by
\begin{equation} \label{eq:disp_volume}
\tD^{P_{0 \rightarrow c}} = \textstyle \sigma\left(\left[g\left(\tD^L_n, d_n, R_c, t_c, \tK_0, \tK_c\right)\right]^{N-1}_{n=0}\right),
\end{equation}
where $\sigma$ is the channel-wise softmax operation and $[\cdot]$ is the channel-wise concatenation operation. $\tD^L$ is the disparity logit volume which can be soft-maxed, quantized, and sum-reduced to obtain the final disparity estimate
\begin{equation} \label{eq:disp_pladenet}
\hat{\tD} = \textstyle \sum_{n=0}^{N-1} d_n \sigma(\tD^L)_n.
\end{equation}
In \cite{falnet}, the authors defined $d_n = d_{max}e^{(n/N-1)\ln{d_{max}/d_{min}}}$ as a fixed exponential quantization. Such quantization scheme distributes the sampling levels more uniformly in the depth domain as described by the gray line in Fig. \ref{fig:adaptive_quant}. However, this assumes that all pixels in the input image will follow the same distribution. Properly allocating `depth bins' to each pixel can distribute probabilities more intensively in the region of interest, generating a much more accurate disparity map. For example, a more \enquote{linear} distribution could benefit the depth estimation of close-by objects by assigning more sampling levels to larger disparities as shown by the green line in Fig. \ref{fig:adaptive_quant}. On the other hand, a steeper distribution should benefit the depth estimation of distant objects, as more bins are assigned to smaller disparities, as depicted by the red line in Fig. \ref{fig:adaptive_quant}. 

The quantization curves depicted in Fig. \ref{fig:adaptive_quant} belong, in fact, to the same family of curves. This family of curves is controlled by the adaptive quantization parameter $\beta$, which we propose to incorporate in a per-pixel ($\vp = (u,v)$) manner in our new adaptive quantization scheme, as given by
\begin{equation} \label{eq:ada_exp_disc}
    \vd(\vp) = \left[d_{max}e^{\ln (d_{max}/d_{min}) \left((n/N)^{\beta(\vp)}-1\right)
    }\right]^N_{n=0}.
\end{equation}
Then, our final inverse depth estimate is given by $\hat{\tD} = \textstyle \vd \cdot \sigma(\tD^L) \inlineeqno$, where $\cdot$ denotes the channel-wise dot product. Note that in \cite{falnet}, $d_n$ is a scalar value, while $\vd$ is a tensor in our proposed adaptive quantization. It is also worth noting that the hypothetical inverse depth planes in Eqs. \ref{eq:synth_free} and \ref{eq:disp_volume}, that were originally single scalar values in \cite{falnet}, become hypothetical inverse depth maps with different values per pixel location by our adaptive quantization scheme. Fig. \ref{fig:adaptive_quant} provides a visualization of the learned adaptive quantization. The adaptive parameter $\beta$ controls the per-pixel disparity quantization curve as depicted in the center of Fig. \ref{fig:adaptive_quant}. As can be observed, $\beta$ effectively assigns the most suitable quantization curve for far and close by object pixels. $\sigma(\tD^L)$ is depicted to the right of Fig. \ref{fig:adaptive_quant} and effectively discretizes the estimated disparity in several levels.

The previous work \cite{adabins} also explored adaptive discretization of depth for the fully supervised case of SVDE. However, while they generate $N$ independently adaptive depth bins globally shared across all pixel locations, we propose a per-pixel quantization so that each output pixel has its own quantization curve. One weakness of AdaBins \cite{adabins} is that it requires GT depths to supervise the bin centers, making it not suitable for the task of interest in the present manuscript, self-supervised learning of SVDE. Furthermore, note that our adaptive quantization scheme not only serves as a disparity representation, but we can also efficiently perform forward warping with it.


\subsection{Network Architecture}
\label{sec:network_arch}
As depicted in Fig. \ref{fig:pipelines}-b, learning SVDE requires at least two branches: one for disparity and the other for relative pose estimation. The first is carried out by our Adaptive QUAntization Network, referred to as `AQUANet', with takes as input a single view $\tI_0$ and its corresponding positional information $(\tU, \tV)$ and produces the disparity logit volume $\tD^L$ and the per-pixel adaptive quantization parameter $\beta$. For AQUANet, we adopted the network backbone in \cite{pladenet}, which is a simple auto-encoder with skip connections that incorporates neural positional encoding (NPE). NPE \cite{pladenet} simply maps pixel coordinates $(u,v)$ to a higher dimensional space via a fully connected network to be processed by the CNN's encoder.

The PoseNet \cite{sfmlearner} is a convolutional encoder that takes target and source images as input and maps them into relative 6 DOF (degrees of freedom) camera extrinsics ($x$, $y$, and $z$ rotations and translations).

\subsection{Depth Cue-Preserving Data Augmentation}
\label{sec:data_aug}

We propose the use of random resize and random crop data augmentations to train our SVDE network, which are well-known in fully supervised learning. In \cite{falnet,pladenet}, these data augmentations have already been exploited when learning from stereo. However, interestingly and to the best of our knowledge, random resize and crop are commonly avoided in most (if not all) previous works that learn from videos \cite{sfmlearner,monodepth2,packing3d,semguide}. Applying these data augmentations can be achieved if the changes in scale and cropping coordinates are properly reflected into the camera focal length and principal points, respectively, with a few caveats. 

Firstly, the pose estimation network requires to \enquote{look} at full-images, with little regard to the image resolutions, to be able to reason about global camera motion. This is not an issue as the final outputs of the PoseNet are just six values per reference camera. So we can feed low-res full images to the PoseNet. 

Secondly, we observed that learning can be controlled in different ways according to the result of the random resizing and cropping operations. Re-scaling the estimated camera translations relative to the random resizing factors can guide the SVDE network to generate disparity estimates that are:
\begin{itemize}[leftmargin=*,noitemsep,topsep=0.5pt]
\vspace*{1mm}
\item \textbf{Invariant to image resolutions.} This is the case when the estimated camera translations are not re-scaled by the random scale factor. 

\item \textbf{Inversely proportional.} If the estimated camera translations are multiplied by the resizing scale factor, the SVDE network will learn to generate smaller-valued disparities as the resolution of the image grows. This is, \textit{the larger the world objects are, the farther away they are}. While this might seem to be the most reasonable approach, it breaks one of the most important cues in SVDE: \textit{relative object size} - the closer the object is, the bigger its relative size.

\item \textbf{Directly proportional.} Dividing the camera translations by the resizing scale factor leads the SVDE model to generate disparity values that scale with image resolutions. This seems to be counter-intuitive, as it could suggest that the larger the world is, the closer the objects are. However, this approach effectively augments the dataset without breaking the relative object size cue.
\end{itemize}
\vspace*{1mm}
In addition to adopting disparities that scale directly proportional to image resolutions, we also incorporate random horizontal flip, random gamma, random brightness, and random color shifts into our training as in \cite{monodepth1,falnet}.

\subsection{Computation of Robust Moving Object Masks - Positional Information is All You Need}
\label{sec:moving_object}
The neural positional encoding (NPE) proposed in the PLADE-Net \cite{pladenet} was known to be simple yet effective in learning monocular depth from stereo images as it provides the SVDE network with the means of understanding the relative location of a random crop with respect to the full image. 

\begin{figure*}[t]
  \centering 
  \includegraphics[width=0.99\textwidth]{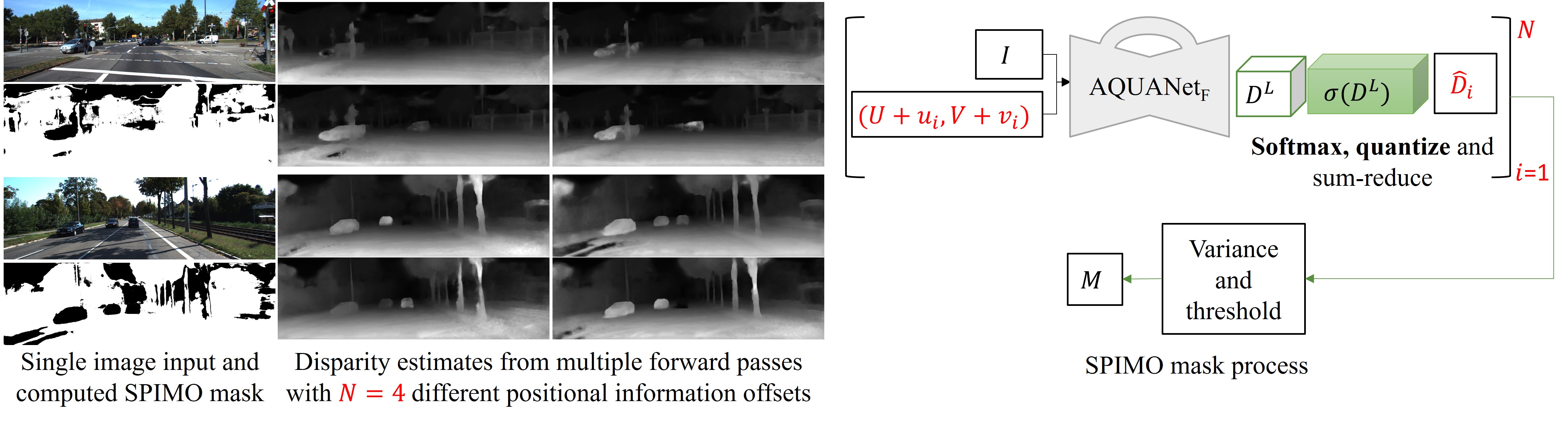}
  \vspace*{-4mm}
  \caption{Computation process of our proposed Shifted Positional Information Moving Object (SPIMO) masks.}
  \label{fig:moving_mask}
  \vspace*{-5mm}
\end{figure*}

However, when learning from monocular videos, the NPE causes the network to memorize the potential locations and appearances of independently moving objects (e.g., cars) in the scene. This might not be desirable for the generalization of SVDE as it can overly accentuate the effects of moving objects into the depth estimates, thus resulting in incorrect depths. However, by exploiting this behavior, we can \textit{fool} a fully trained network to recover the geometries of dynamically moving objects. We \textit{fool} the network by feeding the shifted positional information into the NPE, which enforces the network to switch from an \textit{overfitting} mode into a \textit{generalization} mode, by producing different depth estimates on the regions that were exposed to a weaker 3D self-supervision (moving objects, highly homogeneous regions, etc.) as shown in the bottom part of Fig. \ref{fig:moving_mask}. Note that \textit{positional information is all you need}, as it is used as a mechanism to `trick' the network to recover the missing geometries.

The recovered geometries could be used to learn explainability masks \cite{sfmlearner}, but we observed that learning to predict such masks is hard for the networks, not allowing the loss functions to completely ignore the pixels that belong to moving objects. In this paper, for the first time, we propose another way to exploit the recovered geometries obtained with the shifted positional information. We propose to measure the disparity variance per-pixel, and based on that, classify it as a likely moving or static object pixel in our shifted positional information moving object (SPIMO) masks. Interestingly, our proposed approach allows for obtaining a \textit{likely} moving object mask from a single image rather than a moving object mask from multiple images. To this extent, we propose a learning scheme with \textit{positional information is all you need} to train an SVDE network from videos. Based on this approach, neither an extra network for estimating explainability masks nor a dedicated network for computing optical flows is needed. Instead, the positional information is alone enough to tackle the ill-posed problem of learning SVDE from videos. 

The process for obtaining our variance-based SPIMO masks is depicted on the top of Fig. \ref{fig:moving_mask} and is as follows. First, we build a perturbed depth volume $\tD^v$ with multiple network passes under shifted positional information, as:
\begin{equation} \label{eq:depth_volume}
\tD^v = \left[\text{AQUANet}_F(\tI, (\tU+u_i, \tV+v_i))^{-1}\right]^N_{i=1},
\end{equation}
where each of its $N$ channels is a depth estimate with positional offsets $(u_i, v_i)$. The $\text{AQUANet}_F$ denotes a fixed copy of our network after the first training stage, as shown in top of Fig. \ref{fig:moving_mask}. With normalized positional coordinates (image center at $(0,0)$, top-left corner at $(-1,-1)$), we empirically set $u=[0,0.5,-0.5,0]$ and $v=[0,0,0,-0.25]$ positional offsets. Note that we use depth instead of inverse depth in $\tD^v$ because the variations between close and far away estimates are more evident in depth units.
The \textit{likely} moving object mask is then given by applying a threshold $\gamma$ on the normalized channel-wise variance as given by
 \begin{equation} \label{eq:M_0}
 \tM = 
 \left \{
   \begin{tabular}{cc}
   $1$ & $if\ ({\sum_{i=1}^{N}(\tD^v_i - \bar{\tD}^v)^2})/({(\bar{\tD}^v)^2(N-1)}) < \gamma$ \\
   $0$ & $o.w.$ \\
   \end{tabular}
 \right.,
 \end{equation}
which is similar to the index of dispersion. We empirically set $\gamma = 3\%$ for all our experiments as it was observed to drop most moving object pixels while maintaining rigid object pixels.

\subsection{Boosting as self-supervision for moving objects}
\label{sec:boosting}
We observed that the quality of moving object depth estimates (e.g., cars, bikers) produced by the second training stage with SPIMO masks was much lower than that of the static objects (e.g., roads, trees, traffic signs). This is due to the network not being exposed to those regions during training and such masked regions being only penalized by the smoothness loss. 
To prevent this, we propose to use a watered-down version of the boosting techniques in \cite{boosting_midas,dbooster} to self-supervise the moving objects in the masked regions. Define the disparity estimates from a previous epoch copy of the AQUANet under training at full, reduced and augmented resolutions as 
\begin{align*}
\hat{\tD}_F &= \text{AQUANet}'(\tI_0, (\tU, \tV)),\\
\hat{\tD}_{F\downarrow} &= \tfrac{4}{3}H(\text{AQUANet}'(H(\tI_0, \tfrac{3}{4}), H((\tU, \tV), \tfrac{3}{4})), \tfrac{4}{3}), \text{and}\\
\hat{\tD}_{F\uparrow} &= \tfrac{4}{5}H(\text{AQUANet}'(H(\tI_0, \tfrac{5}{4}), H((\tU, \tV), \tfrac{4}{5})), \tfrac{4}{5}),
\end{align*}
respectively. Where $H(\cdot)$ denotes the resizing operation. Then, the boosted disparity can be computed by a selective combination of $\hat{\tD}_F$, $\hat{\tD}_{F\downarrow}$, and $\hat{\tD}_{F\uparrow}$, as
\begin{equation} \label{eq:boost_disp}
\tD^* = (\hat{\tD}_F + \bar{\tD}_{F}\odot\hat{\tD}_{F\downarrow} + (1 - \bar{\tD}_{F}^2)\odot\hat{\tD}_{F\uparrow}) / (1 + \bar{\tD}_{F} + (1 + \bar{\tD}_{F}^2))
\end{equation}
where $\bar{\tD}_F$ is the normalized mean disparity.
\begin{equation} \label{eq:boost_disp_mean}
\bar{\tD}_F=(\hat{\tD}_F+\hat{\tD}_{F\downarrow}+\hat{\tD}_{F\uparrow})/max(\hat{\tD}_F+\hat{\tD}_{F\downarrow}+\hat{\tD}_{F\uparrow}).
\end{equation}

Eq. \ref{eq:boost_disp} effectively blends the three estimates from the fixed network, assigning more weights on the upscaled estimates for faraway objects and more weights on the downscaled estimates for the close-by objects, producing a slightly better depth estimate, as described in \cite{boosting_midas,dbooster}.

\begin{figure*}[t]
  \centering 
  \includegraphics[width=0.99\textwidth]{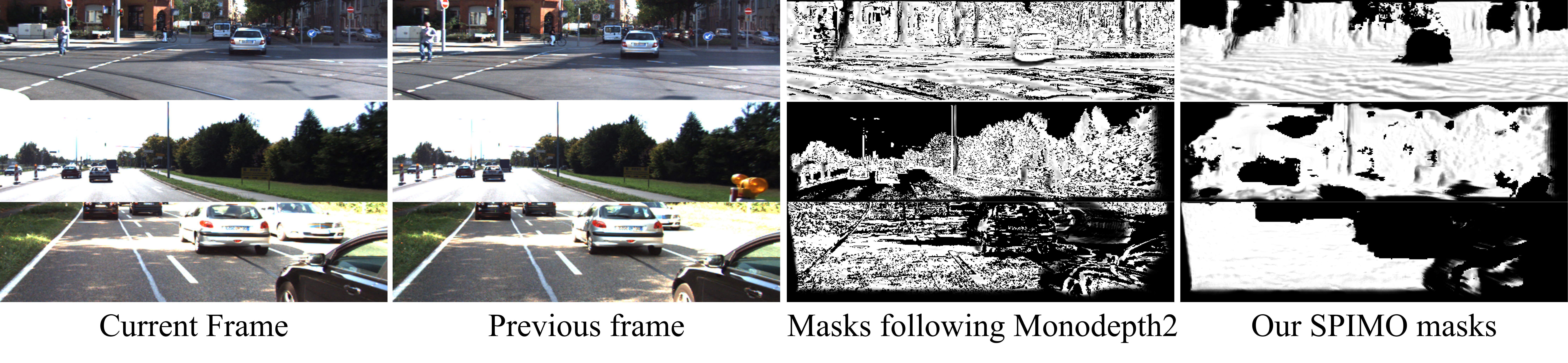}
  \vspace*{-4mm}
  \caption{Visual comparison between our shifted positional information moving object (SPIMO) masks and those computed following Monodepth2 \cite{monodepth2}.}
  \label{fig:vs_mask}
  \vspace*{-2mm}
\end{figure*}

\subsection{Loss Functions}
\label{sec:loss_funct}

In the first training stage, we learn from all non-occluded image pixels with a combination of occlusion-aware synthesis $l^{o}_{s}$ and edge-aware disparity smoothness $l_{ds}$ losses, as given by $l_{st1} = l^o_{s} + \alpha_{ds}l_{ds}$, where $\alpha_{ds} = 0.1$ following \cite{monodepth1,packing3d,falnet}. 

In the second stage, we train the network from scratch with an occlusion and moving object-aware synthesis loss $l^{OM}_{s}$, $l_{ds}$, and a boosting loss $l_{b}$, as given by
\begin{equation} \label{eq:l_st2}
l_{st2} = l^{om}_{s} + \alpha_{ds}l_{ds} + \alpha_{b}l_{b},
\end{equation}
where $\alpha_{b} = 0.1$ is set empirically. Our synthesis loss, $l^{om}_{s} = 0.5l^{om}_{s_1} + 0.5l^{om}_{s_{-1}}$, averages the contribution of each previous and next reference views.

\vspace{2mm}
\noindent
\textbf{Occlusion / moving object-aware synthesis loss.}
Given the combination of occlusion and SPIMO masks $\tW_1 = \tO_1 \odot \tM_1 \odot \tM_{0 \rightarrow 1}$, where $\tM_{0 \rightarrow 1}$ is the target frame's mask projected into the reference camera, this term is defined as
\begin{equation} \label{eq:l_om_syn}
l^{om}_{s_1} = ||\tW_1 \odot (\tI'_1 - \tI_1)||_1 + \alpha_p\textstyle\sum_{l=1}^{3} ||\phi^l(\tI'_1)-\phi^l(\tI^w_1)||^2_2,
\end{equation}
where $\tI^w_1=(1 - \tW_1) \odot \tI'_1 + \tW_1 \odot \tI_1$ is the GT view with moving objects replaced by those in $\tI'_1$. The weight $\alpha_p = 0.01$ is set empirically to balance the contribution between the L1 and perceptual losses \cite{perceptual} and $\phi^l(\cdot)$ denotes the $l^{th}$ maxpool layer of a pre-trained VGG19\cite{vgg}. Note that $\tO_1$ is obtained from $\tD^L$ following the method in \cite{falnet}. $l^{o}_{s}$ in $l_{st1}$ is obtained by just setting $\tW = \tO$ in $l^{om}_{s}$.


\vspace{2mm}
\noindent
\textbf{Boosting loss.}
This loss builds on the previous works \cite{boosting_midas,dbooster} and provides refined disparity maps $\tD^*$ to self-supervise the moving object regions, and is given by
\begin{equation} \label{eq:boosting_loss}
l_b = \tfrac{1}{max(\tD^*)}||(1 - \tM_0) \odot (\hat{\tD} - \tD^*)||_1.
\end{equation}
Note that the boosting loss is only applied to the pixel regions that are likely moving independently, normalized by the maximum disparity value $max(\tD^*)$.


\section{Experiments and Results}
We trained AQUANet with our novel pipeline in two stages for 110 epochs in each stage. For the first 5 epochs, we kick-start the PoseNet with a `naive disparity map'. This map is a vertical gradient image where the first and last image rows are assigned zeros and ones, respectively. We implemented our pipeline with PyTorch and trained it on an NVIDIA A100 GPU with a batch size of 8. We optimized our networks using the default Adam \cite{adam} optimizer with an initial learning rate of 0.0001. The learning rate is halved at epochs [60, 80, 100]. As mentioned in Section \ref{sec:data_aug}, we applied random cropping (for all datasets) of 192$\times$640 and random resizing from 0.5 to 1.5. We set the disparity discretization levels as 33, which well fits our adaptive quantization scheme.

\subsection{Datasets}
 
\textbf{KITTI (K)\cite{kitti2012}.}
To compare our results with a wide range of previous works, we utilized the Eigen train and test splits \cite{eigen} of the KITTI dataset. The train split contains 22,600 384$\times$1208 image sequences captured at 10Hz. The test split\cite{eigen} contains 697 and 652 images with projected sparse LiDAR ground truths (which we cap at 80\textit{m}), respectively in its original \cite{eigen} and improved versions \cite{kitti_official}. The latter contains multi-view aggregated higher-quality depths.

\noindent
\textbf{CityScapes (CS)\cite{cityscapes}.}
We utilize the train folder of the CS dataset, which contains $\sim$3K 384$\times$1208 images, each surrounded by 29 frames captured at 17Hz and adding up to $\sim$80K training sequences. To ensure large enough motion in the sequences, we skip every other frame, reducing the frame rate to half. 

\begin{table*}[t]
    \scriptsize
    \centering
    \caption{Ablation studies of our AQUANet on KITTI\cite{kitti2012}. Metrics defined in \cite{eigen}. Error metrics are \colorbox{c_lowbest}{the lower the better} and accuracy metrics are \colorbox{c_highbest}{the higher the better}.}
    \vspace*{-3mm}
    \setlength{\tabcolsep}{3pt}
    \begin{tabular}{clccccccc}
\hline
Expmt. & Methods & abs rel\cellcolor{c_lowbest} & sq rel\cellcolor{c_lowbest} & rmse\cellcolor{c_lowbest} & rmse$_{log}$\cellcolor{c_lowbest}  & $\delta^1$\cellcolor{c_highbest} & $a^2$\cellcolor{c_highbest} & $a^3$\cellcolor{c_highbest} \\ 
\hline

1&AQUANet (1st) & 0.120 & 1.457 & 4.286 & 0.167 & 0.891 & 0.970 & 0.986 \\
&AQUANet (1st-stat) & 0.080 & 0.287 & 2.858 & 0.114 & 0.936 & 0.992 & 0.999 \\
\hline

2 &AQUANet (with \cite{monodepth2} masks) & 0.113 & 0.820 & 4.236 & 0.167 & 0.881 & 0.970 & 0.989 \\
& DIRNet-BW & 0.098 & 0.579 & 3.910 & 0.149 & 0.900 & 0.976 & 0.992 \\
& vPLADE-Net  & 0.084 & 0.366 & 3.240 & 0.124 & 0.930 & 0.987 & 0.997 \\
& AQUANet &  0.080 & 0.328 & 3.095 & 0.118 & 0.934 & 0.989 & 0.997 \\
\hline

3 & AQUANet (no aug.) & 0.098 & 0.532 & 3.923 & 0.146 & 0.902 & 0.978 & 0.993 \\
& AQUANet (inverse)   & 0.091 & 0.417 & 3.446 & 0.131 & 0.918 & 0.986 & 0.996 \\
& AQUANet (invariant) & 0.087 & 0.385 & 3.288 & 0.126 & 0.925 & 0.987 & 0.997 \\
\hline

4 & AQUANet($a_b=0.1$) & 0.078 & 0.320 & 3.117 & 0.117 & 0.935 & 0.989 & 0.997 \\
&AQUANet(full) & 0.077 & 0.324 & 3.032 & 0.115 & 0.938 & 0.989 & 0.997 \\
\hline

\end{tabular}
    \label{tab:ablation_extra}
    \vspace*{-2mm}
\end{table*}

\begin{figure*}[t]
  \centering 
  \includegraphics[width=1.0\textwidth]{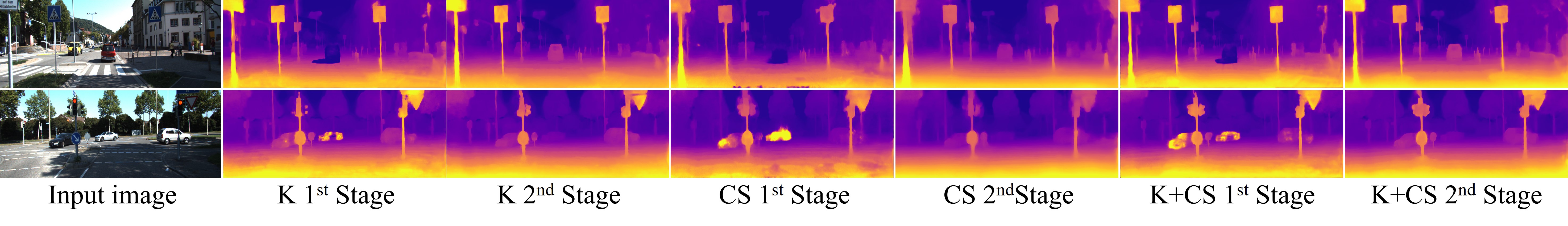}
  \vspace*{-9mm}
  \caption{AQUANet results on KITTI images for 1st and 2nd training stages.}
  \label{fig:res_data}
  \vspace*{-1mm}
\end{figure*}

\vspace*{-4mm}
\subsection{Ablation Studies}
Table \ref{tab:ablation_extra} shows the results of extensive ablation studies on the KITTI dataset to inspect the effectiveness of each component in our method. 

\noindent
\textbf{Experiment 1 (Expmt. 1) in Table \ref{tab:ablation_extra}.} We validate our SPIMO masks by training our AQUANet for SVDE without masking any moving object, denoted as AQUANet(1st). The various KITTI metrics \cite{eigen} are poor due to the moving objects for which infinite or incorrect depths are often predicted, as depicted in the 2nd column of Fig. \ref{fig:res_data}. We then measure the performance of AQUANet(1st) on the non-moving objects only, guided by SPIMO masks, denoted as AQUANet(1st-stat). The better metrics of AQUANet(1st-stat) imply that our SPIMO masks can effectively remove the scene's moving objects.

\begin{figure*}[t]
  \centering 
  \includegraphics[width=1.0\textwidth]{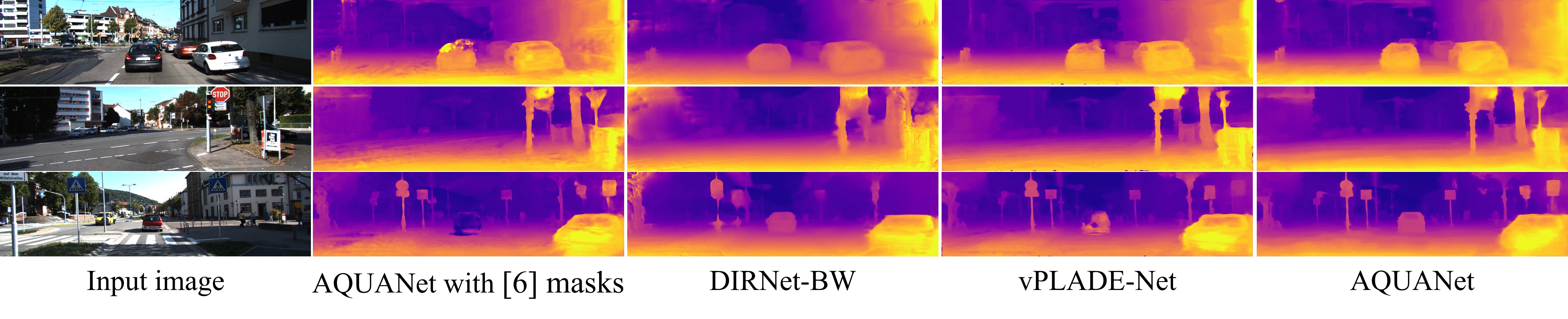}
  \vspace*{-9mm}
  \caption{Qualitative comparison among different network architectures.}
  \label{fig:ablation_1}
  \vspace*{-2mm}
\end{figure*}

\begin{figure*}[t]
  \centering 
  \includegraphics[width=0.99\textwidth]{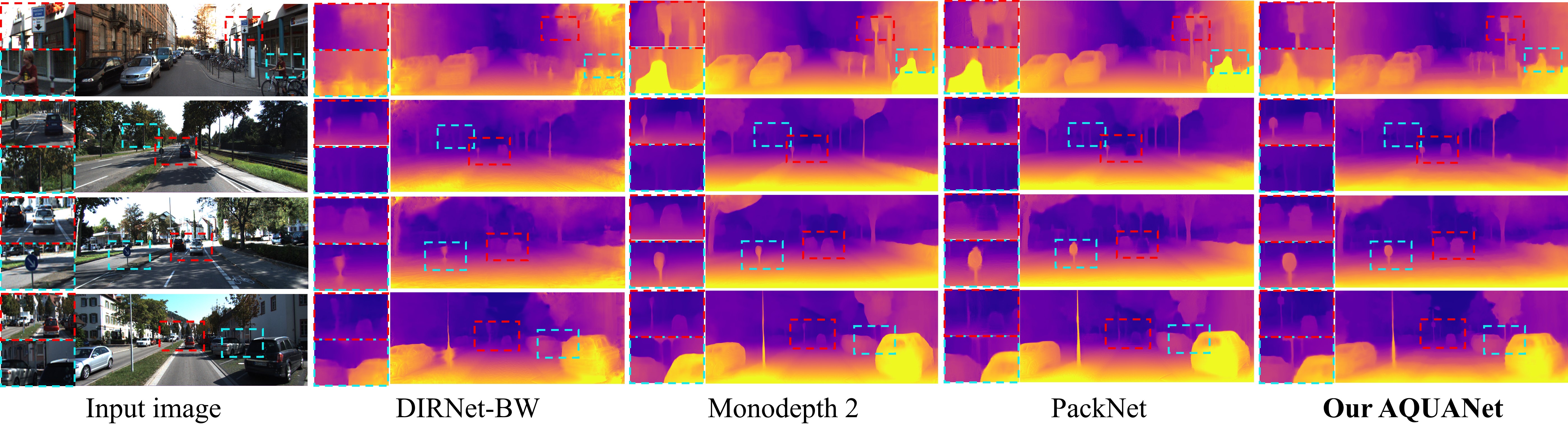}
  \vspace*{-4mm}
  \caption{Qualitative comparisons on KITTI. Our AQUANet consistently estimates more detailed depths and is robust against moving objects thanks to our SPIMO masks.}
  \label{fig:res_eigen}
  \vspace*{-5mm}
\end{figure*}

\noindent
\textbf{Expmt. 2.} We inspect our second training stage on different architectures with similar network backbones. Firstly, we train our AQUANet with moving object masks from \cite{monodepth2}. As can be noted in Table \ref{tab:ablation_extra} and Fig. \ref{fig:ablation_1}, the masks in \cite{monodepth2} are not enough to entirely remove the moving objects in the scene, forcing the AQUANet with \cite{monodepth2} masks to generate holes in the estimated disparity maps and considerably lower quantitative performance. Fig. \ref{fig:vs_mask} depicts our SPIMO masks for various KITTI \cite{kitti2012} sequences and their corresponding auto-masks following \cite{monodepth2}. As observed, SPIMO masks are much less sparse and removes the independently moving objects better, while the photometric assumptions in \cite{monodepth2} do not always manage to remove the moving objects in the scene.

Secondly, the DIRNet-BW, which learns using backward warping based loss functions (like \cite{monodepth2}), but with our SPIMO masks, manages to estimate full depth maps but struggles in estimating plausible depths for the thin objects in the scene. This is reflected in the improved metrics in Expmt. 2 of Table \ref{tab:ablation_extra} and qualitative results of Fig. \ref{fig:ablation_1}. Thirdly, the vPLADE-Net, a video extension of \cite{pladenet} with our SPIMO masks but without their expensive `Matting Laplacian loss,' yields improved metrics and more detailed depth maps but presents disparity artifacts near the image borders (such as small holes) and incorrect depths around the scene's objects boundaries, as shown in Table \ref{tab:ablation_extra} and Fig. \ref{fig:ablation_1}. Finally, our AQUANet, with SPIMO masks and pixel-wise adaptive quantization, remarkably obtains the best results in Expmt. 2, both quantitatively and qualitatively, as depicted in Table \ref{tab:ablation_extra} and Fig. \ref{fig:ablation_1}.

\noindent
\textbf{Expmt. 3.} We ablate the effects of our proposed data augmentation protocol, for AQUANets that are trained without random resizing and cropping augmentations, denoted as AQUANet (no aug.), AQUANets that learn to estimate disparity values that don't scale with image resolutions, denoted as AQUANet (invariant), and AQUANets that learn to estimate disparities that scale inversely proportional to image resolutions, denoted as AQUANet (inverse).
These networks achieve much lower performance than our AQUANet that learns disparity values that scale proportional to the image resolutions as shown in Table \ref{tab:ablation_extra}.
Note that, AQUANet (no aug.) is limited to predicting depth maps at half resolution, as it was not exposed to randomly resized and cropped patches during training.

\noindent
\textbf{Expmt. 4.} We ablate the effects of boosting into our training. Our AQUANet with boosting ($a_b=0.1$) achieves improved results in most metrics. However, we observed that even better results could be achieved if the boosted depths $\tD^*$ are incorporated into Eq. \ref{eq:depth_volume} in the SPIMO mask computation process, denoted as AQUANet(full). The considerable differences between the first and second training stages on the different datasets can be observed in Fig. \ref{fig:res_data}.

\subsection{Results}
\textbf{KITTI.} A quantitative comparison between our method and existing SOTA methods that learn from videos is presented in Table \ref{tab:kitti}. Results for methods that learn from stereo or depth GT are also given for reference. Our method always achieves the best sq rel and RMSE metrics in comparison with the previous self-supervised methods when trained with (K) only and with (K+CS) for both the original and the improved Eigen test splits \cite{eigen}. For the improved Eigen test split (which contains much denser depths and is therefore more reliable), our AQUANet outperforms the SOTA method of PackNet \cite{packing3d} in 4 out of 7 metrics by a considerable margin and is comparable for the other three metrics even though the PackNet has 8.5 times more parameters than ours. Our superior quantitative results are supported by qualitatively more consistent depth estimates, as shown in Fig. \ref{fig:res_eigen}. 
    
\noindent
\textbf{CityScapes.} Firstly, we demonstrate that our method benefits from additional data, as shown in the (K+CS) results of Table \ref{tab:kitti}. Secondly, to test the generality of our method, we train our models on (CS) and test them on (K). The consistent improvements are observed in Table \ref{tab:kitti} from the first training stage (RMSE: 5.396) to the second stage (RMSE: 4.686) regardless of the training data, which implies that our method generalizes well. Additionally, 
as shown in the second column of Fig. \ref{fig:res_data}, our AQUANet, even with no KITTI images during training, predicts the moving car in the KITTI scene in the first row with infinite depths and the moving cars in the second row with implausible close-by depths. These experimental results imply that CNN's trained for the SVDE task from videos do not memorize moving objects but rather learn their appearances and potential locations. \textit{More experimental results are provided in the supplementary materials}.



\begin{table*}[t]
    \scriptsize
    \centering
    \caption{Comparison to Existing SVDE Methods on the KITTI Eigen Split \cite{eigen}. DoF: depth of field supervision. D: Fully-supervised. V, V$_{\text{se}}$, S, and S$_{\text{SGM}}$: Self-supervised from video, video+semantics, stereo, and stereo+SGM. Models train on KITTI (K) or CityScapes (CS). PP: post-processing. Par: \# of params. V models use median scaling.}
    \vspace*{-2mm}
    \setlength{\tabcolsep}{2.5pt}
\begin{tabular}{>{\cellcolor{c_data}}cclcccccccccc}
\hline
& Ref & Methods & Sup & Data & Par & abs rel\cellcolor{c_lowbest}  & sq rel\cellcolor{c_lowbest}  & rmse\cellcolor{c_lowbest}  & log$_{rmse}$\cellcolor{c_lowbest}  & $a^1$\cellcolor{c_highbest} & $a^2$\cellcolor{c_highbest} & $a^3$\cellcolor{c_highbest} \\ 
\cline{2-13}

&\cite{defocus} & Gur \etal & DoF & K & - & \underline{0.110} & \underline{0.666} & \textbf{4.186} & \textbf{0.168} & \underline{0.880} & \textbf{0.966} & \textbf{0.988} \\ 
&\cite{singleviewstereo} & Luo \etal & D+S  & K & - & \textbf{0.094} & \textbf{0.626} & \underline{4.252} & \underline{0.177} & \textbf{0.891} & \underline{0.965} & \underline{0.984} \\	
\cline{2-13}

&\cite{depth_hints}    & DepthHints(P)  & S$_{\text{SGM}}$ & K & 35 & 0.096 & 0.710 & 4.393 & 0.185 & 0.890 & 0.962 & 0.981 \\
&\cite{pladenet} & PLADE-Net & S & K+CS & 15 & \textbf{0.090} & \underline{0.577} & \textbf{3.880} & \textbf{0.170} & \textbf{0.903} & \textbf{0.968} & \textbf{0.985} \\
\cline{2-13}

&\cite{monodepth2}   & Monodepth2 & V & K & 14 & 0.115 & 0.882 & 4.701 & 0.190 & 0.879 & 0.961 & 0.982 \\	
&\cite{packing3d}    & PackNet & V & K & 120 & 0.107 & 0.802 & 4.538 & 0.186 & 0.889 & 0.962 & 0.981 \\
&\cite{hr_depth}   & HR-Depth & V & K & 14 & \underline{0.104} & 0.727 & 4.410 & \underline{0.179} & 0.894 & \textbf{0.966} & \textbf{0.984} \\	
&- & AQUANet & V & K & 14 & 0.115 & \textbf{0.656} &  \textbf{4.251} & 0.186 & 0.875 & 0.959 & \underline{0.983} \\

&\cite{packing3d}    & PackNet & V & CS$\rightarrow$K & 120 & \underline{0.104} & 0.758 & 4.386 & 0.182 & \underline{0.895} & 0.964 & 0.982 \\ 
&\cite{semguide}     & Guizilini \etal & V+Se & CS$\rightarrow$K & 140 & \textbf{0.100} & 0.761 & 4.270 & \textbf{0.175} & \textbf{0.902} & \underline{0.965} & 0.982 \\

\multirow{-12}{*}{\rotatebox[origin=c]{90}{Original Eigen Test Split\cite{kitti2012}}}
&- & AQUANet & V & K+CS & 14 & 0.112 & \underline{0.669} & \underline{4.263} & 0.182 & 0.882 & 0.963 & \textbf{0.984} \\
\hline

\hline
&\cite{dorn}       & DORN      & D & K & 51 & 0.072 & 0.307 & \textbf{2.727} & 0.120  & 0.932 & 0.984 & 0.995 \\
&\cite{vnormal}    & Yin \etal & D & K & 113 & \textbf{0.072} & -     & 3.258 & \textbf{0.117} & \textbf{0.938} & \textbf{0.990} & \textbf{0.998} \\
\cline{2-13}

&\cite{depth_hints} & DepthHints(P) & S$_{\text{SGM}}$ & K & 35 & 0.074 & 0.364 & 3.202 & 0.114 & 0.936 & 0.989 & \underline{0.997} \\
&\cite{pladenet} & PLADE-Net & S & K+CS & 15 & \textbf{0.066} & \textbf{0.263} & \textbf{2.726} & \textbf{0.102} & \textbf{0.949} & \textbf{0.992} & \textbf{0.998} \\
\cline{2-13}

&\cite{monodepth2} & Monodepth2 & V & K & 14 & 0.092 & 0.536 & 3.749 & 0.135 & 0.916 & 0.984 & 0.995 \\
&\cite{packing3d}  & PackNet & V & K & 120 & 0.078 & 0.420 & 3.485 & 0.121 & 0.931 & 0.986 & 0.996 \\
&- & AQUANet($1^{st}$) & V & K & 14 & 0.120 & 1.457 & 4.286 & 0.167 & 0.891 & 0.970 & 0.986 \\
&- & AQUANet & V & K & 14 & 0.079 &  \underline{0.324} &  \underline{3.032} & 0.115 & 0.938 & 0.989 & 0.997 \\

&\cite{packing3d}  & PackNet & V & CS$\rightarrow$K & 120 & \textbf{0.071} & 0.359 & 3.153 & \textbf{0.109} & \textbf{0.944} &  \underline{0.990} & \underline{0.997} \\ 
&- & AQUANet($1^{st}$) & V & K+CS & 14 & 0.114 & 1.171 & 4.055 & 0.162 & 0.898 & 0.975 & 0.988 \\
&- & AQUANet & V & K+CS & 14 &  \underline{0.076} & \textbf{0.301} & \textbf{2.921} &  \underline{0.111} &  \underline{0.943} & \textbf{0.991} & \textbf{0.998} \\
&- & AQUANet($1^{st}$) & V & CS & 14 & 0.179 & 1.881 & 5.396 & 0.240 & 0.790 & 0.933 & 0.969 \\
&- & AQUANet & V & CS & 14 & 0.131 & 0.786 & 4.686 & 0.186 & 0.837 & 0.962 & 0.988 \\ 
\cline{2-13}
\multirow{-14}{*}{\rotatebox[origin=c]{90}{Improved Test Split\cite{kitti_official}}}
&- & AQUANet(P) & V & K+CS & 14 & 0.073 & 0.279 & 2.813 & 0.107 & 0.946 & 0.992 & 0.998 \\

\hline
\end{tabular}
    \label{tab:kitti}
    \vspace*{-4mm}
\end{table*}

\section{Conclusions}
We have proposed the use of positional information for the first time to aid in self-supervised learning of SVDE from videos. We showed that an SVDE network implicitly learns the potential locations and appearances of moving objects. This behavior can be exploited to compute robust moving object masks, which we call SPIMO masks. We used these masks in our novel learning pipeline, making it possible to learn SVDE from randomly cropped patches by conveniently handling the changes in disparity scales. Additionally, we presented adaptive quantization, utilized by our AQUANet, which in conjunction with our novel learning pipeline, yields the SOTA results for the self-supervised learning of SVDE from videos.

\clearpage
%
%
\bibliographystyle{splncs04}
\bibliography{main_bib}
\end{document}